\documentclass[tikz]{article}

\usepackage{PRIMEarxiv}

\usepackage[utf8]{inputenc} 
\usepackage[T1]{fontenc}    
\usepackage{hyperref}       
\usepackage{url}            
\usepackage{booktabs}       
\usepackage{amsfonts}       
\usepackage{nicefrac}       
\usepackage{microtype}      
\usepackage{lipsum}
\usepackage{fancyhdr}       
\usepackage{graphicx}       
\usepackage{csquotes}
\usepackage{mathtools} 
\usepackage{booktabs} 
\usepackage{algorithm}
\usepackage[noend]{algorithmic}
\usepackage{multirow}
\usepackage{caption}
\usepackage{subcaption}
\usepackage{amssymb}
\usepackage{graphicx}
\usepackage{xfrac}
\usepackage{xcolor}
\usepackage{ulem}
\usepackage{microtype}
\usepackage{hyperref}
\usepackage[acronym]{glossaries}
\usepackage{cleveref}

\title{
    Investigating the Impact of Large-Scale Pre-training on Nutritional Content Estimation from 2D Images
}

\author{
Michele Andrade$^1$, Guilherme A. L. Silva$^1$, Valéria Santos$^1$,\\ 
\textbf{Gladston Moreira$^1$ and Eduardo Luz}$^1$ \\
 $^1$Computing Department, Universidade Federal de Ouro Preto, Ouro
Preto, 35402-136, Minas Gerais, Brazil.\\
    \texttt{eduluz@ufop.edu.br}
}

\begin{document}
\maketitle

\begin{abstract}
    Estimating the nutritional content of food from images is a critical task with significant implications for health and dietary monitoring.
This is challenging, especially when relying solely on 2D images, due to the variability in food presentation, lighting, and the inherent difficulty in inferring volume and mass without depth information. Furthermore, reproducibility in this domain is hampered by the reliance of state-of-the-art methods on proprietary datasets for large-scale pre-training.
In this paper, we investigate the impact of large-scale pre-training datasets on the performance of deep learning models for nutritional estimation using only 2D images. We fine-tune and evaluate Vision Transformer (ViT) models pre-trained on two large public datasets, ImageNet and COYO, comparing their performance against baseline CNN models (InceptionV2 and ResNet-50) and a state-of-the-art method pre-trained on the proprietary JFT-300M dataset.
We conduct extensive experiments on the Nutrition5k dataset, a large-scale collection of real-world food plates with high-precision nutritional annotations. Our evaluation using Mean Absolute Error (MAE) and Mean Absolute Percentage Error (MAE\%) reveals that models pre-trained on JFT-300M significantly outperform those pre-trained on public datasets. Unexpectedly, the model pre-trained on the massive COYO dataset performs worse than the model pre-trained on ImageNet for this specific regression task, refuting our initial hypothesis.
Our analysis provides quantitative evidence highlighting the critical role of pre-training dataset characteristics, including scale, domain relevance, and curation quality, for effective transfer learning in 2D nutritional estimation.
\end{abstract}


\section{Introduction}
\label{sec:introduction}

Estimating the nutritional content of food solely from images presents a significant challenge at the intersection of computer vision and nutrition science. The growing global focus on health, dietary management, and chronic disease prevention has amplified the need for accessible and accurate tools to track nutrient intake. The ability to determine calories, carbohydrates, proteins, and fats from a simple photograph of a meal would revolutionize personal health monitoring, dietary studies, and nutritional support systems.

This task is inherently complex despite its potential, particularly when relying on standard 2D image data from ubiquitous devices like smartphone cameras. Key difficulties include the immense variability in food presentation (lighting, angle, plating), the challenge of distinguishing individual ingredients within mixed dishes, and critically, the ambiguity in estimating serving size and volume from a single 2D perspective. Traditional methods often require manual input or supplementary data (e.g., food scales, measuring tools), limiting scalability and user convenience.

Recent research has attempted to tackle these challenges through diverse approaches. Datasets like Nutrition5k \cite{thames2021nutrition5k}, pic2kcal \cite{ruede2021multi}, ISIA Food-500 \cite{min2020isia}, and Food2K \cite{min2023large} have emerged, providing large collections of food images, often with associated nutritional information and ingredient lists. Some studies have explored the use of depth sensors \cite{thames2021nutrition5k, naritomi2021hungry} or explicit volume estimation techniques \cite{meyers2015im2calories} to mitigate the limitations of 2D data. Deep learning models, particularly Convolutional Neural Networks (CNNs), have shown promise in recognizing food items and even predicting nutritional values directly from images \cite{ruede2021multi, chen2016deep, yunus2018framework, chen2020study}. However, many state-of-the-art approaches, such as the baseline presented with the Nutrition5k dataset \cite{thames2021nutrition5k}, achieve their top performance by leveraging models pre-trained on massive, often proprietary, datasets like JFT-300M \cite{sun2017revisiting}.

This reliance on inaccessible pre-training data poses a significant challenge to the reproducibility of research in the community. It restricts widespread adoption, further development, and fair comparison of new methods. Motivated by this, our work focuses on investigating the effectiveness of publicly available large-scale pre-training datasets for nutritional estimation from 2D images, specifically exploring the potential of the Vision Transformer (ViT) architecture \cite{dosovitskiy2020image} due to its strong performance on large-scale vision tasks.

This study aims to answer the following research questions:
\begin{itemize}
    \item Can Vision Transformer models effectively estimate nutritional content (calories, mass, macronutrients) from 2D food images when fine-tuned on a specific food dataset?
    \item How does pre-training a ViT model on a massive, publicly available multimodal dataset (COYO) compare to pre-training on a widely used visual dataset (ImageNet) for this regression task?
    \item Can pre-training on public datasets (ImageNet or COYO) achieve performance comparable to models pre-trained on large proprietary datasets (JFT-300M)?
\end{itemize}
Based on the scale of the COYO dataset, which is closer in size to JFT-300M than ImageNet, we hypothesize that pre-training on COYO would yield superior results compared to pre-training on ImageNet and potentially bridge the performance gap with the JFT-300M baseline for 2D nutritional estimation.

We formulate the problem as a multi-task regression task predicting total calories, mass, fat, carbohydrates, and proteins. We implement and evaluate ViT models pre-trained on ImageNet and COYO, fine-tuning them on the Nutrition5k dataset \cite{thames2021nutrition5k}. We compare their performance using MAE and MAE\% against standard CNN baselines (InceptionV2, ResNet-50) trained under similar conditions and the reported results of the state-of-the-art baseline from \cite{thames2021nutrition5k}.

Our main contributions are:
\begin{itemize}
    \item Adaptation and evaluation of the Vision Transformer architecture for nutritional content regression from 2D food images;
    \item Empirical investigation into the impact of pre-training on large-scale public datasets (ImageNet and COYO) for this specific regression problem, comparing them directly;
    \item Quantitative comparison of ViT performance against established CNN architectures and a state-of-the-art baseline pre-trained on a proprietary dataset, highlighting the challenges of achieving competitive results without access to such data;
    \item Providing insights into which characteristics of large pre-training datasets (scale, domain, curation) appear most critical for successful transfer learning in 2D nutritional estimation.
\end{itemize}

\section{Related Work}
\label{sec:related_work}

Developing robust food recognition and nutritional estimation methods relies heavily on the availability of large, diverse, and accurately annotated datasets. Several efforts have focused on collecting food image datasets, often with varying levels of annotation detail. Datasets such as pic2kcal \cite{ruede2021multi} and ISIA Food-500 \cite{min2020isia} were constructed by scraping images and associated information from websites, providing large-scale collections. Pic2kcal includes recipes and nutritional details, while ISIA Food-500 focuses on fine-grained food category recognition across 500 categories. Food2K \cite{min2023large} further expands this by collecting over 1 million images for 2000 categories, demonstrating broad applicability in tasks like recognition and retrieval. While these datasets are valuable for food category recognition, their reliance on crowdsourced or web-scraped nutritional data can introduce inaccuracies, and they often lack precise information about serving sizes or ingredient weights for specific dishes depicted in images.

Addressing the crucial challenge of portion size estimation, the Nutrition5k dataset \cite{thames2021nutrition5k} provides a unique resource by collecting 5,000 real-world food plates, along with accompanying videos, depth images, and precise component weights, thereby enabling highly accurate nutritional annotations. This structured data collection approach contrasts with web-scraping and provides a more reliable ground truth for nutritional estimation. Similarly, Naritomi et al. \cite{naritomi2021hungry} explored the use of 3D scanning to create 3D models of food items and plates, aiding in volume estimation and highlighting the importance of 3D information. Our work utilizes the Nutrition5k dataset, focusing specifically on its 2D components to investigate models applicable in scenarios lacking depth sensors.

Various deep learning architectures have been applied to food recognition and nutritional tasks. CNNs, such as VGG, Inception, and ResNet, have been widely used for food image classification \cite{yunus2018framework} and as backbones for nutritional prediction models \cite{thames2021nutrition5k, ruede2021multi, chen2020study}. Multi-task learning approaches, where a single model simultaneously predicts multiple attributes like ingredients, categories, and nutritional values, have shown benefits by exploiting correlations between tasks \cite{ruede2021multi, chen2020study}. For instance, Ruede et al. \cite{ruede2021multi} used a multi-task setup with CNNs on pic2kcal to predict calories and macronutrients. Thames et al. \cite{thames2021nutrition5k} employed a multi-task regression head on an InceptionV2 base for calorie, mass, and macronutrient prediction on Nutrition5k.

More recently, Vision Transformers (ViT) \cite{dosovitskiy2020image}, inspired by NLP success, have demonstrated state-of-the-art performance in image recognition when pre-trained on massive datasets. Min et al. \cite{min2023large} leveraged transformer-like attention mechanisms in their ``deep progressive region enhancement network'' for food recognition and retrieval on Food2K. While transformers show promise in vision, their application to nutritional regression from images, and the impact of different large-scale pre-training sources on this specific task, remains less explored. Critically, the superior performance of the Nutrition5k baseline \cite{thames2021nutrition5k} is attributed to pre-training on the proprietary JFT-300M dataset, which is inaccessible for most researchers.

Our work directly compares established CNNs (InceptionV2, ResNet-50) with the Vision Transformer for 2D nutritional estimation on Nutrition5k. Unlike methods relying on 3D data, we focus on the challenging 2D scenario. Our primary contribution lies in empirically evaluating the effectiveness of public large-scale pre-training datasets (ImageNet and COYO) for ViT on this regression task, contrasting their performance and characteristics against the JFT-300M pre-trained baseline to shed light on the reproducibility gap and the factors influencing successful transfer learning in this domain.

\section{Background}
\label{sec:background}

The problem of estimating nutritional content from images is approached using supervised learning techniques, specifically deep neural networks, which are highly effective at learning complex patterns from visual data \cite{lecun2015deep}. Our task is formulated as a regression problem, aiming to predict continuous numerical values (mass, calories, and macronutrients) directly from a 2D image.

\subsection{Problem Setting: Nutritional Estimation as Regression}
\label{sec:problem_setting}

We formalize the problem of nutritional content estimation from a single 2D image as a multi-task regression problem.
Given an input food image \(I \in \mathbb{R}^{H \times W \times C}\), where \(H\), \(W\), and \(C\) are the height, width, and number of channels (e.g., 3 for RGB), the objective is to predict a vector \( \hat{Y} = (\hat{y}_w, \hat{Y}_m, \hat{y}_{cal}) \), where:
\begin{itemize}
    \item \(\hat{y}_w \in \mathbb{R}\) is the estimated total mass (in grams) of the plate;
    \item \(\hat{Y}_m \in \mathbb{R}^k\) is a vector of estimated masses (in grams) for \(k\) specific macronutrients. In this work, \(k=3\) for fat, carbohydrates, and protein, so \(\hat{Y}_m = (\hat{y}_{\text{fat}}, \hat{y}_{\text{carb}}, \hat{y}_{\text{protein}})\);
    \item \(\hat{y}_{cal} \in \mathbb{R}\) is the estimated total caloric content (in kcal) of the plate.
\end{itemize}
The ground truth for a given image \(I_i\) is \(Y_i = (y_i^w, Y_i^m, y_i^{cal})\), derived from measured weights of individual ingredients and their known nutritional values per gram. The task is to learn a mapping function \(f: \mathbb{R}^{H \times W \times C} \to \mathbb{R}^{k+2}\) that minimizes a defined loss function between the predicted values \( \hat{Y}_i = f(I_i) \) and the ground truth \(Y_i\) over a dataset \(D = \{I_i, Y_i\}_{i=1}^N\).

\subsection{Deep Learning Models}
\label{sec:deep_learning_models}

We investigate three distinct deep learning architectures commonly used in computer vision and adapted for regression: InceptionV2, ResNet-50, and the Vision Transformer (ViT).

\subsubsection{InceptionV2}
InceptionV2 \cite{szegedy2016rethinking} improves upon the original Inception architecture by factorizing large convolutions into smaller or asymmetric ones to enhance computational efficiency while maintaining representation power. It employs parallel branches with different filter sizes within inception modules.

\subsubsection{ResNet-50}
ResNet-50 \cite{he2016deep} is a 50-layer network utilizing residual connections (\enquote{skip connections}) to alleviate the degradation problem in deep networks. These connections facilitate gradient flow, enabling the effective training of models that are significantly deeper than standard feedforward architectures.

\subsubsection{Vision Transformer (ViT)}
\begin{figure}[htbp]
    \centering
    \includegraphics[width=0.99\columnwidth]{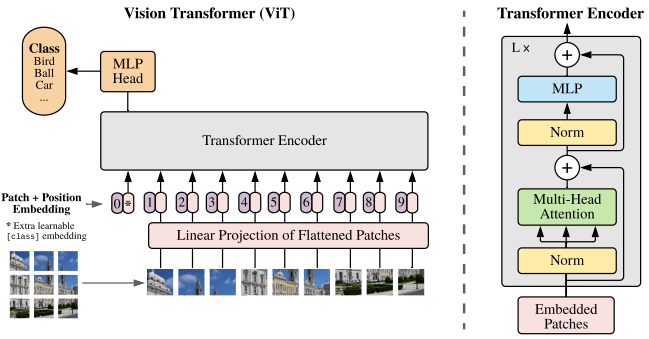}
    \caption{Vision Transformer (ViT) Architecture. (Source: \cite{dosovitskiy2020image})}
    \label{fig:vit_architecture}
\end{figure}

The Vision Transformer (ViT) \cite{dosovitskiy2020image} adapts the Transformer architecture \cite{vaswani2017attention} for image processing. It treats an image as a sequence of patches, processes them using self-attention mechanisms in a Transformer encoder, and lacks the strong inductive biases of CNNs, relying heavily on large-scale pre-training data to achieve high performance. As illustrated in \Cref{fig:vit_architecture}, input images are divided into patched, embedded, and processed by a standard Transformer encoder.

\subsection{Multi-task Learning Head}
\label{sec:multitask_head}

Following the approach in \cite{thames2021nutrition5k}, we employ a multi-task learning head on top of the backbone CNN (InceptionV2, ResNet-50) or ViT model. This head shares initial fully connected layers across tasks before splitting into task-specific final layers. For CNNs, the output feature map from an intermediate layer (specifically, `mixed5c' for InceptionV2 and the layer before global average pooling for ResNet-50) is spatially pooled (3x3 average pooling with stride 2). For ViT, the output embedding of the classification token from the last layer is used directly.

A sequence of two shared fully connected (FC) layers, each with 4096 dimensions, processes the pooled features and token embedding. Subsequently, separate task-specific FC layers (also with 4096 dimensions) are used for mass, calories, and a combined head for macronutrients. Each task has a final FC layer mapping to its respective output dimension (1 for mass, calories, 3 for macronutrients).

The overall loss function is a summation of the losses for each task, trained jointly:
\begin{equation}
\begin{aligned}
\mathcal{L}(I, Y|W) &= \mathcal{L}_{m}(I, Y^{m}|W) + \mathcal{L}_{c}(I, y^{\text{cal}}|W) + \mathcal{L}_{w}(I, y^{w}|W) \\
\end{aligned}
\end{equation}
where $W$ represents the model parameters, $I$ is the input image, and $Y = (y^w, Y^m, y^{cal})$ are the ground truth labels. For the CNN experiments compared to the baseline, we use Mean Absolute Error (MAE) as the loss for each task, consistent with \cite{thames2021nutrition5k}:
\begin{equation}
\begin{aligned}
\mathcal{L}_{m}(I, Y^{m}|W) &= \frac{1}{|M|} \sum_{j \in M} | \hat{y}_{j}^{m}(I|W) - y_{j}^{m} | \\
\mathcal{L}_{c}(I, y^{\text{cal}}|W) &= | \hat{y}^{\text{cal}}(I|W) - y^{\text{cal}} | \\
\mathcal{L}_{w}(I, y^{w}|W) &= | \hat{y}^{w}(I|W) - y^{w} |
\end{aligned}
\end{equation}
where \(M = \{ \text{fat, carbohydrate, protein} \}\), $\hat{y}(I|W)$ denotes the model's prediction for input $I$ with parameters $W$, and $y$ denotes the ground truth label. For ViT experiments, we used Mean Squared Error (MSE) as the loss function for each task, a common choice for regression problems that was found to work well empirically during initial testing. The overall loss function for ViT training thus becomes:
\begin{equation}
\begin{aligned}
\mathcal{L}_{ViT}(I, Y|W) &= \text{MSE}( \hat{Y}_m(I|W), Y_m ) + \text{MSE}( \hat{y}_{\text{cal}}(I|W), y_{\text{cal}} ) \\ &+ \text{MSE}( \hat{y}_{w}(I|W), y_{w} )
\end{aligned}
\end{equation}
where $\text{MSE}(a, b) = \frac{1}{N} \sum_{i=1}^N (a_i - b_i)^2$ for vectors $a, b$ of size $N$, or $(a-b)^2$ for scalars.

\begin{figure}[htbp]
    \centering
    \includegraphics[width=0.9\columnwidth]{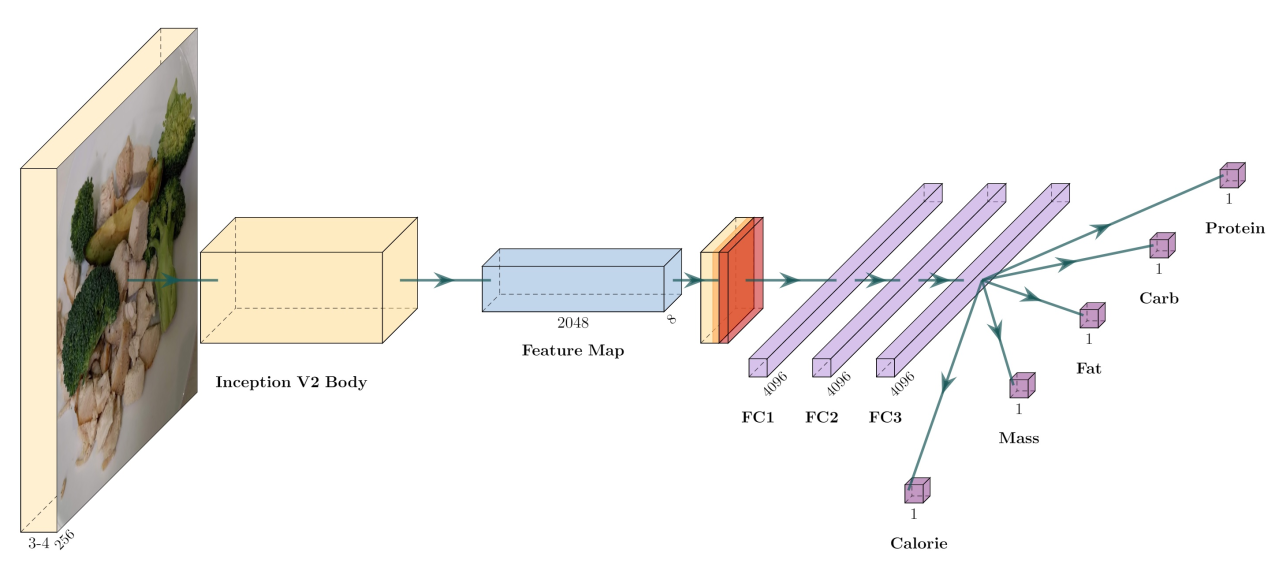}
    \caption{Overall approach for nutritional content estimation from a 2D food image using a deep learning model. (Adapted from \cite{thames2021nutrition5k})}
    \label{fig:overall_approach}
\end{figure}

\Cref{fig:overall_approach} illustrates the overall approach of estimating nutritional content from a 2D image using a deep learning model with a multi-task approach.

\section{Experimental Setup}
\label{sec:experimental_setup}

\subsection{Dataset}
\begin{figure}[htbp]
    \centering
    \includegraphics[width=0.9\columnwidth]{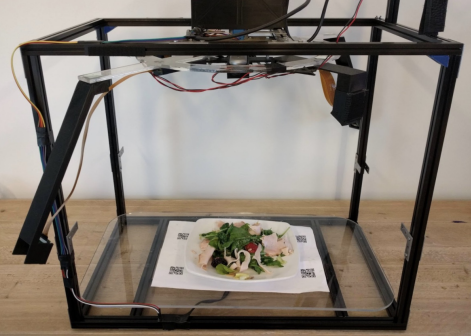}
    \caption{Data collection equipment used for the Nutrition5k dataset. (Source: \cite{thames2021nutrition5k})}
    \label{fig:data_collection}
\end{figure}

Our experiments are conducted using the Nutrition5k dataset \cite{thames2021nutrition5k}. This dataset was specifically designed for nutritional understanding from images, providing high-precision annotations. It comprises approximately 5,000 unique real-world food plates, built from over 250 distinct ingredients. The data collection process involved a custom sensor system that weighed and scanned each plate component individually (\Cref{fig:data_collection}). Each plate is annotated with total weight, weights for individual components, and derived nutritional values (calories, fat, carbs, protein) calculated from USDA data. The dataset also includes RGB videos and aerial RGB-D images for many plates.

For this study, which focuses on the challenging task of nutritional estimation from 2D images only, we utilize the RGB video data provided in Nutrition5k. We extract static frames from these videos to simulate standard 2D image input. Specifically, we sample one frame every five frames from the recorded videos associated with each plate.

The dataset is split following the original paper's protocol: 90\% of the plates are used for training, and the remaining 10\% are held out for testing. This split is performed at the plate level to ensure that no images from the same plate appear in both the training and testing sets, thereby providing a realistic evaluation of generalization. The dataset exhibits variability in food types, portions, and quantities, reflecting real-world culinary diversity, as illustrated by the most common ingredients by mass (\Cref{tab:common_ingredients}).


\begin{table}[h!]
    \centering
        \caption{Frequency of ingredient usage (in thousands)}
    \begin{tabular}{c|c}
        \toprule
        \parbox{110pt}{\centering \textbf{Ingredient}} & \parbox{110pt}{\centering \textbf{Value (k)}} \\
        \midrule
        Egg whites & 47.8 \\
        Chicken & 47.7 \\
        Scrambled eggs & 40.3 \\
        Olives & 37.3 \\
        Cauliflower & 29.6 \\
        Broccoli & 29.3 \\
        Lemon & 28.0 \\
        Berries & 24.3 \\
        Fish & 21.3 \\
        Apple & 19.4 \\
        Cheese pizza & 19.2 \\
        Sweet potato & 18.3 \\
        Carrot & 18.2 \\
        Pineapple & 17.6 \\
        Caesar salad & 17.2 \\
        Cantaloupe & 16.3 \\
        Cherry tomatoes & 15.7 \\
        Beef & 15.5 \\
        Pork & 15.3 \\
        Salmon & 14.8 \\
        Sausage & 14.0 \\
        Roasted potatoes & 13.7 \\
        Asparagus & 13.6 \\
        Brussels sprouts & 13.3 \\
        White rice & 13.2 \\
        Mixed greens & 12.9 \\
        Squash & 11.6 \\
        Corn on the cob & 10.4 \\
        Cucumbers & 10.2 \\
        Watermelon & 10.2 \\
        \bottomrule
    \end{tabular}
    \label{tab:common_ingredients}
\end{table}

\subsection{Models and Training Details}
We implement and train three types of models: InceptionV2, ResNet-50, and Vision Transformer (ViT).

CNN Baselines (InceptionV2 and ResNet-50): We implemented InceptionV2 based on the description in \cite{thames2021nutrition5k} and ResNet-50 \cite{he2016deep} using standard architectures. Both models were initialized with weights pre-trained on the ImageNet dataset \cite{ILSVRC15}. Input images were resized and center-cropped to $256 \times 256$ pixels. Training utilized the RMSprop optimizer with a learning rate of $1 \times 10^{-4}$, momentum 0.9, decay 0.9, and epsilon 1.0. The multi-task head described in \Cref{sec:multitask_head} was attached, and models were trained using the L1 loss function for each task.

Vision Transformer (ViT): We used the CLIP ViT architecture available from the Hugging Face Transformers library. The pre-trained classification head was removed, and a linear regression head was added on top of the output embedding of the classification token from the last Transformer layer. This head maps to 5 output values: total mass, total fat, total carbs, total protein, and total calories.

We explored two different pre-training sources for the ViT model:
\begin{itemize}
    \item ViT pre-trained on ImageNet: A ViT model pre-trained on the standard ImageNet-21k dataset.
    \item ViT pre-trained on COYO: A ViT model pre-trained on the large-scale COYO-700M dataset \cite{kakaobrain2022coyo-700m}, which contains over 700 million image-text pairs.
\end{itemize}
For fine-tuning on Nutrition5k, both ViT models were trained using the Adam optimizer with an empirically tuned learning rate. The multi-task MSE loss function was used. Training was conducted for two different durations to observe convergence behavior: 60 epochs and 300 epochs. Batch Normalization was applied in the regression head.

\subsection{Evaluation Metrics}
To evaluate the performance of the models in estimating total mass, calories, and individual macronutrient masses (fat, carbs, protein), we use the Mean Absolute Error (MAE) and Mean Absolute Percentage Error (MAE\%).

The Mean Absolute Error (MAE) is defined as:
\begin{equation}
    MAE = \frac{1}{n} \sum_{i=1}^{n} |y_i - \hat{y}_i|
\end{equation}
where $n$ is the number of samples in the test set, $y_i$ is the ground truth value for the $i$-th sample, and $\hat{y}_i$ is the predicted value for the $i$-th sample.

The Mean Absolute Percentage Error (MAE\%) provides a relative measure of error:
\begin{equation}
    MAE\% = \frac{1}{n} \sum_{i=1}^{n} \left| \frac{y_i - \hat{y}_i}{y_i} \right| \times 100\%
\end{equation}
Note that MAE\% can be unstable or undefined for ground truth values close to zero. However, for the aggregated metrics (total calories, total mass, total macronutrient masses), ground truth values are generally non-zero and substantial.

Lower values for both MAE and MAE\% indicate better model performance.

\section{Results and Discussion}
\label{sec:results}

This section presents the experimental results and discusses their implications, particularly addressing our research questions and hypothesis regarding the impact of pre-training datasets on 2D nutritional estimation.

\subsection{CNN Baselines vs. Nutrition5k Baseline}
\Cref{tab:cnn_comparison} presents the performance of our implemented InceptionV2 and ResNet-50 models, both pre-trained on ImageNet and fine-tuned on Nutrition5k using L1 loss, compared to the reported performance of the Nutrition5k baseline \cite{thames2021nutrition5k}, which used an InceptionV2-based architecture pre-trained on JFT-300M.

\begin{table}[htbp]
    \centering
    \caption{Comparison of MAE and MAE\% for InceptionV2 and ResNet-50 (ImageNet pre-trained) against the Nutrition5k Baseline (JFT-300M pre-trained).}
    \resizebox{\linewidth}{!}{
    \begin{tabular}{lcccccc}
        \toprule
         & \multicolumn{2}{c}{InceptionV2 (ImageNet)} & \multicolumn{2}{c}{ResNet-50 (ImageNet)} & \multicolumn{2}{c}{Baseline (JFT-300M)} \\
        Metric & MAE & MAE \% & MAE & MAE \% & MAE & MAE \% \\
        \midrule
        Total Calories & 140.88 & 55.90 & 117.38 & 46.57 & \textbf{70.6} & \textbf{26.1} \\
        Total Mass & 82.50 & 43.62 & 66.83 & 35.33 & \textbf{40.4} & \textbf{18.8} \\
        Total Fat & 9.93 & 73.00 & 9.02 & 66.32 & \textbf{5.0} & \textbf{34.2} \\
        Total Carb & 11.19 & 58.48 & 9.80 & 51.21 & \textbf{6.1} & \textbf{31.9} \\
        Total Protein & 10.54 & 67.14 & 8.92 & 56.80 & \textbf{5.5} & \textbf{29.5} \\
        \bottomrule
    \end{tabular}}
    \label{tab:cnn_comparison}
\end{table}

\begin{figure}[htbp]
    \centering
    \begin{subfigure}[b]{0.49\textwidth}
        \includegraphics[width=\textwidth]{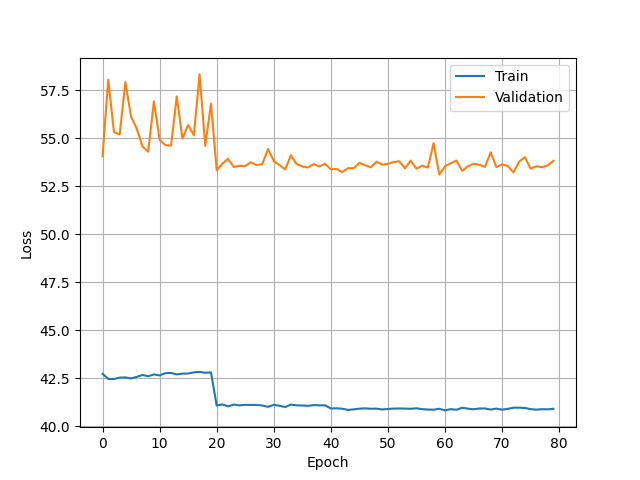}
        \caption{InceptionV2 (L1 loss)}
    \end{subfigure}
    \hfill
    \begin{subfigure}[b]{0.49\textwidth}
        \includegraphics[width=\textwidth]{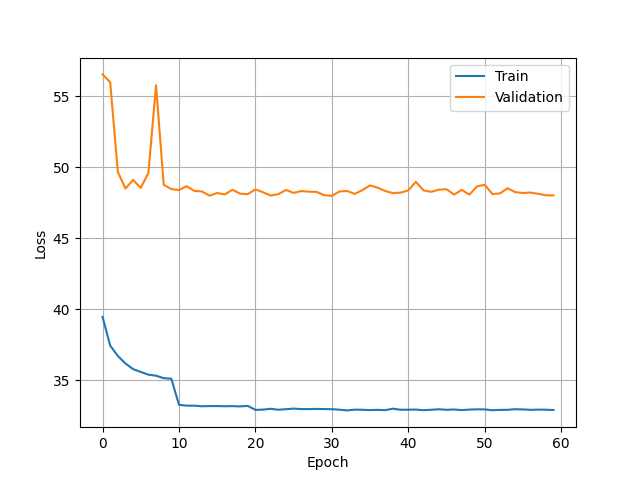}
        \caption{ResNet-50 (L1 loss)}
    \end{subfigure}
    \caption{Training and validation loss curves for InceptionV2 and ResNet-50 models trained with L1 loss.}
    \label{fig:cnn_loss_curves}
\end{figure}

As observed, the ResNet-50 model consistently outperforms the InceptionV2 model across all metrics when both are pre-trained on the ImageNet dataset. This suggests that the ResNet architecture's ability to train deeper models effectively through residual connections provides an advantage for this task, even compared to the InceptionV2 variant used in the baseline. \Cref{fig:cnn_loss_curves} shows the training and validation loss curves, indicating relatively stable convergence for both models, although ResNet-50's performance is better overall.

However, despite ResNet-50's better performance compared to InceptionV2, both models fall significantly short of the Nutrition5k baseline in terms of MAE and MAE\%. The baseline achieves substantially lower errors across all nutritional parameters. This stark difference highlights the impact of the pre-training dataset. The baseline was pre-trained on JFT-300M \cite{sun2017revisiting}, a proprietary dataset with 300 million images and 375 million labels. This massive scale and diversity, encompassing a broad range of visual concepts, provide a much richer starting point for fine-tuning compared to ImageNet (approximately 14 million images). This initial comparison underscores the challenge of replicating state-of-the-art results in vision tasks when the original models leverage inaccessible pre-training data.

\subsection{Vision Transformer (ViT) Experiments}
The results of the ViT experiments are presented, specifically aimed at evaluating the impact of different large-scale public pre-training datasets (ImageNet and COYO) on 2D nutritional estimation, and assessing their effectiveness in approaching the performance of the JFT-300M baseline.

\Cref{tab:vit_comparison_60epochs} and \Cref{tab:vit_comparison_300epochs} show the performance of ViT models pre-trained on COYO and ImageNet, trained for 60 and 300 epochs, respectively, compared to the Nutrition5k baseline.

\begin{table}[htbp]
\tiny
    \centering
    \caption{Comparison of MAE and MAE\% for ViT models (60 Epochs) against the Nutrition5k Baseline.}
    \resizebox{\linewidth}{!}{
    \begin{tabular}{lcccccc}
        \toprule
         & \multicolumn{2}{c}{ViT (COYO)} & \multicolumn{2}{c}{ViT (ImageNet)} & \multicolumn{2}{c}{Baseline (JFT-300M)} \\
        Metric & MAE & MAE \% & MAE & MAE \% & MAE & MAE \% \\
        \midrule
        Total Calories & 144.40 & 57.29 & \textbf{87.99} & \textbf{34.91} & \textbf{70.6} & \textbf{26.1} \\
        Total Mass & 91.98 & 48.63 & \textbf{63.87} & \textbf{33.77} & \textbf{40.4} & \textbf{18.8} \\
        Total Fat & 10.48 & 77.04 & \textbf{7.22} & \textbf{53.07} & \textbf{5.0} & \textbf{34.2} \\
        Total Carb & 12.19 & 63.68 & \textbf{8.28} & \textbf{43.28} & \textbf{6.1} & \textbf{31.9} \\
        Total Protein & 11.47 & 73.03 & \textbf{6.63} & \textbf{42.24} & \textbf{5.5} & \textbf{29.5} \\
        \bottomrule
    \end{tabular}}
    \label{tab:vit_comparison_60epochs}
\end{table}

\begin{table}[htbp]
\tiny
    \centering
    \caption{Comparison of MAE and MAE\% for ViT models (300 Epochs) against the Nutrition5k Baseline.}
    \resizebox{\linewidth}{!}{
    \begin{tabular}{lcccccc}
        \toprule
         & \multicolumn{2}{c}{ViT (COYO)} & \multicolumn{2}{c}{ViT (ImageNet)} & \multicolumn{2}{c}{Baseline (JFT-300M)} \\
        Metric & MAE & MAE \% & MAE & MAE \% & MAE & MAE \% \\
        \midrule
        Total Calories & 136.97 & 54.34 & \textbf{95.29} & \textbf{37.81} & \textbf{70.6} & \textbf{26.1} \\
        Total Mass & 82.23 & 43.48 & \textbf{60.55} & \textbf{32.01} & \textbf{40.4} & \textbf{18.8} \\
        Total Fat & 10.38 & 76.27 & \textbf{7.40} & \textbf{54.43} & \textbf{5.0} & \textbf{34.2} \\
        Total Carb & 11.42 & 59.64 & \textbf{8.29} & \textbf{43.30} & \textbf{6.1} & \textbf{31.9} \\
        Total Protein & 10.21 & 65.05 & \textbf{6.86} & \textbf{43.67} & \textbf{5.5} & \textbf{29.5} \\
        \bottomrule
    \end{tabular}}
    \label{tab:vit_comparison_300epochs}
\end{table}

Consistent with the CNN results, the Nutrition5k baseline pre-trained on JFT-300M remains superior to both ViT models pre-trained on public datasets across all metrics and training durations. This further reinforces the significant advantage conferred by pre-training on extremely large, diverse, and presumably well-curated datasets like JFT-300M.

Crucially, the results for the ViT models pre-trained on COYO and ImageNet directly address our second and third research questions and test our hypothesis. The ViT model pre-trained on ImageNet significantly outperforms the ViT model pre-trained on COYO in nearly all metrics, for both 60 and 300 epochs. For example, the MAE for Total Calories after 300 epochs is 136.97 for COYO ViT vs 95.29 for ImageNet ViT. This refutes our hypothesis that pre-training on COYO would be superior to ImageNet due to its larger size and purported similarity to JFT-300M. The ImageNet pre-trained ViT consistently yields lower errors, demonstrating better transfer learning capability for this task compared to the COYO pre-trained model.

Several factors could explain the poorer performance of the COYO pre-trained ViT compared to ImageNet. While COYO is massive (~700M image-text pairs), it is a multimodal dataset collected from the web, potentially containing significant noise and less structured annotations compared to ImageNet, which is specifically curated for object classification. More importantly, the domain of images in COYO might be less relevant to the specific visual characteristics of food plates compared to the broad range of objects and scenes in ImageNet, which includes many food categories. Although JFT-300M is also web-scale, Sun et al. \cite{sun2017revisiting} found that its 300M images and 375M labels, despite some noise, provide tremendous visual diversity, leading to representations that generalize exceptionally well. Our results suggest that for the task of nutritional estimation from food images, the curated diversity and focus of datasets like ImageNet (or the extreme scale and breadth of JFT-300M) are more beneficial than the sheer size and multimodal nature of COYO.

\begin{figure}[!ht]
    \centering
    \includegraphics[width=\textwidth]{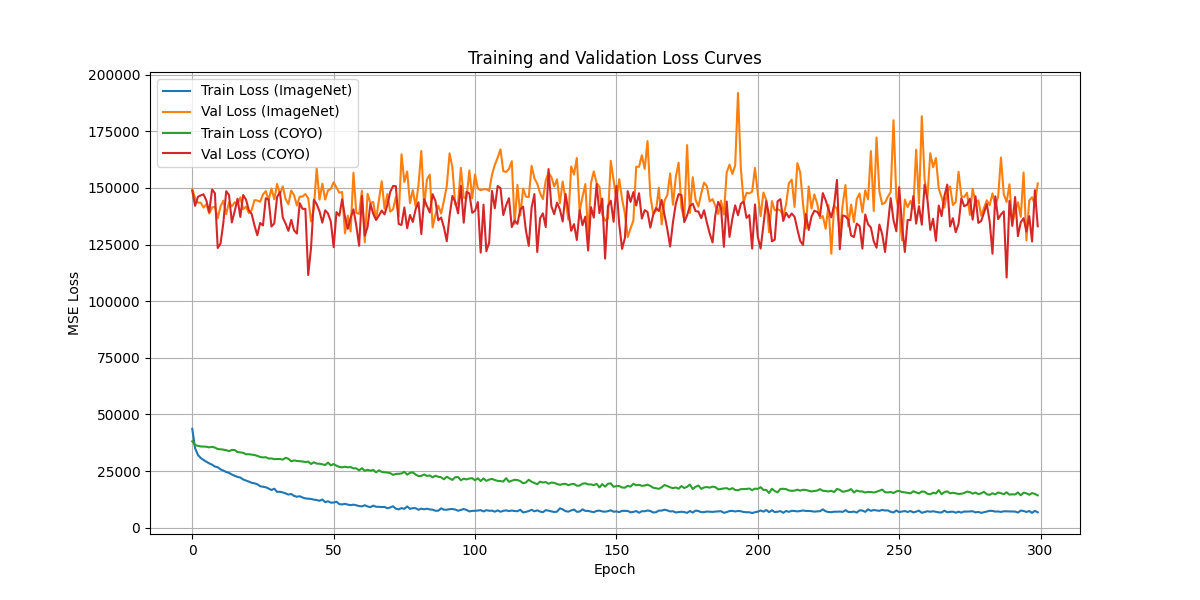}
    \caption{Training and validation loss curves for ViT models pre-trained on COYO and ImageNet (300 Epochs).}
    \label{fig:vit_loss_curves}
\end{figure}

Increasing the training duration from 60 to 300 epochs generally improved the performance of both ViT models, with MAE values decreasing for most metrics (e.g., Calories MAE for COYO: 144.40 at 60 epochs vs 136.97 at 300 epochs; for ImageNet: 87.99 at 60 epochs vs 95.29 at 300 epochs - Note: ImageNet calories MAE slightly increased at 300 epochs, while mass/macro MAE decreased. This oscillation is visible in loss curves). \Cref{fig:vit_loss_curves} illustrates the training and validation loss curves for the 300-epoch runs. The ImageNet ViT exhibits a lower training loss and somewhat more stable validation loss compared to the COYO ViT, indicating better learning on the training data and slightly improved generalization, although both models show some volatility in validation loss, which is typical when fine-tuning large models on smaller datasets. While increased training time helps, it does not bridge the gap to the JFT-300M baseline performance.

\subsection{Discussion on Research Questions}
Regarding our first research question, ViT models, when fine-tuned, can indeed estimate nutritional content from 2D food images, achieving performance significantly better than random but still considerably lower than the JFT-300M baseline. Their effectiveness is heavily dependent on the pre-training data.
For the second question, pre-training on ImageNet resulted in demonstrably better performance for this regression task compared to pre-training on COYO, contrary to our hypothesis.
And addressing the third question, models pre-trained on public datasets (ImageNet and COYO), even the massive COYO, were unable to match the performance of the baseline pre-trained on the proprietary JFT-300M dataset. This highlights the significant advantage that large-scale, possibly more curated and domain-relevant (in terms of visual diversity) proprietary data provides, posing a major hurdle for reproducibility in this field. The results quantitatively demonstrate that achieving state-of-the-art performance requires access to data resources that are not universally available.

The source code and experimental results are available at \url{https://github.com/michele-andrade/Nutricao-Inteligente}.

\section{Conclusion}
\label{sec:conclusion}

In this work, we investigated the problem of estimating nutritional content from 2D food images, focusing on the impact of large-scale pre-training datasets on model performance.

We implemented and evaluated Vision Transformer models pre-trained on the public ImageNet and COYO datasets, comparing them against CNN baselines (InceptionV2, ResNet-50) and a state-of-the-art baseline pre-trained on the proprietary JFT-300M dataset, using the Nutrition5k dataset.

Our experiments yielded clear insights. Firstly, while ViT models can be effectively fine-tuned for nutritional regression, their performance is strongly tied to the pre-training data. Secondly, contrary to our initial hypothesis, pre-training on the massive COYO dataset did not surpass pre-training on ImageNet for this specific task; the ImageNet pre-trained ViT consistently performed better. Finally, and most significantly, neither of the models pre-trained on public datasets could match the performance of the JFT-300M pre-trained baseline.

These findings quantitatively demonstrate that the scale and, more importantly, the specific characteristics (domain relevance, curation quality) of pre-training data are paramount for achieving state-of-the-art results in 2D nutritional estimation. The performance gap observed with the JFT-300M baseline underscores the reproducibility challenge faced by researchers without access to such proprietary resources. Our results suggest that simply using any large public dataset may not be sufficient; the nature of the visual concepts learned during pre-training matters critically for transfer learning to this complex regression task.

\section*{Acknowledgment}
This work was supported by the Conselho Nacional de Desenvolvimento Científico e Tecnológico (CNPq, grants 308400/2022-4, 307151/2022-0), Coordenação de Aperfeiçoamento de Pessoal de Nível Superior (CAPES - grant 001), Fundação de Amparo à Pesquisa do Estado de Minas Gerais (FAPEMIG, grant APQ-01647-22). We also thank the Universidade Federal de Ouro Preto (UFOP) for their support.





\bibliographystyle{main}  
\bibliography{main}

\end{document}